# IMAGE AESTHETIC EVALUATION USING PARALLELED DEEP CONVOLUTION NEURAL NETWORK

*Guo Lihua, Li Fudi*

School of Electronic and Information Engineering, South China University of Technology, Guangzhou 510640, China

## ABSTRACT

Image aesthetic evaluation has attracted much attention in recent years. Image aesthetic evaluation methods heavily depend on the effective aesthetic feature. Traditional methods always extract hand-crafted features. However, these hand-crafted features are always designed to adapt particular datasets, and extraction of them needs special design. Rather than extracting hand-crafted features, an automatically learn of aesthetic features based on deep convolutional neural network (DCNN) is first adopt in this paper. As we all know, when the training dataset is given, the DCNN architecture with high complexity may meet the over-fitting problem. On the other side, the DCNN architecture with low complexity would not efficiently extract effective features. For these reasons, we further propose a paralleled convolutional neural network (PDCNN) with multi-level structures to automatically adapt to the training dataset. Experimental results show that our proposed PDCNN architecture achieves better performance than other traditional methods.

*Kerwords— Image Aesthetic Evaluation, Deep Convolution Neural Network (DCNN), Paralleled Convolutional Neural Network (PDCNN)*

## 1. INTRODUCTION

Image aesthetic evaluation aims to classify photos into high quality or low quality from the perspective of human. As photos shown in Fig 1, most people tend to prefer pictures on the top row since these pictures seem to be more attractive. With the improvement of image aesthetic evaluation system, more and more applications will appear in computer vision area. In the image retrieval system, the aesthetic quality of an image may be an important factor when designing the ranking algorithms. We can also take the image aesthetic quality into account when using the image management system.

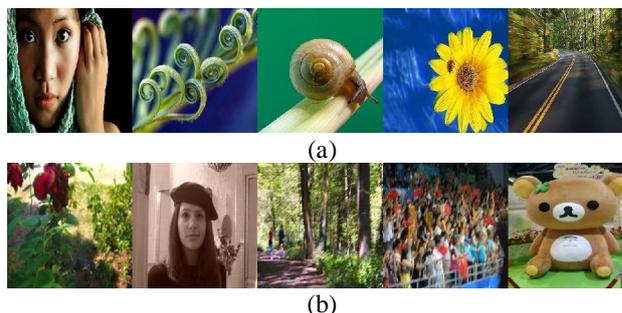

Figure 1. Most people tend to prefer pictures above in (a), in other words, pictures in (a) are of higher quality than those in (b).

Image Aesthetic Evaluation was first put forward by Tong et al. [1], in which they design a "black box" model to classify photos into professional photos and snapshots. Recently, lots of image aesthetic evaluation methods have been proposed. Method [2] proposed some image visual features to evaluate image aesthetic, including exposure, contrast, colorfulness, spatial distribution and color distribution of the photographs. Ke et al. [3] designed high level semantic features to measure the perceptual differences. The highlight of Ke' method was that it gave detailed analysis on several perceptual criteria of aesthetic evaluation, and the aesthetic features were designed based on these criteria. These methods based on low and high level features both performed well, however, they had the limitation that they mainly extracted features from an entire image. In order to overcome this limitation, Luo[4] and Wong[5] both proposed methods of extracting high level semantic features from the subject region and the salient region of an image, which proved to be more effective. In some cases, the extraction of subject region and salient region may be fail, Guo et al. [15] proposed a method to fuse the visual features and semantic features. To obtain more general features instead of hand-crafted features, Luca [11] proposed a method of using general image descriptors, such as BOV, FV and SIFT, which outperforms previous methods by a significant margin.

Though traditional methods have achieved competitive performance, they needed extracting some hand-crafted features which are mainly designed for a particular dataset to gain good performance, so they may not adapt to other datasets. Take feature extraction method in [6] for example, the regional feature named face combined was specially designed for the category of human, and the global feature named scene composition was only designed for the category of night. Unlike hand-crafted feature extraction methods, the feature learning method is a very popular research topic, and has attracted great attention in image processing area. DCNN is one of the feature learning methods, and has achieved great success in solving many computer vision problems, for example, hand-written digit recognition [7], object recognition [8], image classification [9] and so on. However, few attempts[12] [13] have been made to apply DCNN into image aesthetic evaluation. For these reasons, we introduce DCNN to automatically learn aesthetic features rather than designing hand-crafted features. The most common problem of DCNN is over-fitting and under-fitting problems. A lot of strategies have been taken to avoid these problems, such as dropout[10], conjugate gradient[17], validation[18]. Our strategy is to parallel DCNNs with multi-level structures. A DCNN architecture of high complexity can suit for a large-scale dataset, and a DCNN architecture of low-complexity is suitable for a small-scale dataset. So paralleling architectures of different complexity is able to improve the fitness for different scale datasets.

The main contribution of our paper is that we not only apply DCNN to image aesthetic evaluation, but also propose PDCNN architectures to overcome over-fitting and under-fitting problems. The rest of our paper is organized as follows. Section 2 describes the traditional DCNN architecture we used and gives a description on our PDCNN architecture. Section 3 demonstrates the effectiveness of our method by extensive experiments. Our concluding remarks are given in Section 4.

## 2. OUR ALGORITHM

In this section, we will first give detailed information of the proposed traditional DCNN architecture. Then we will conduct experiments to select the number of convolutional layers of our DCNN architecture, because applying DCNN to image aesthetic evaluation is not a straightforward task, and it is definitely important to select a suitable DCNN architecture.

### 3.1 Traditional DCNN architecture

The traditional DCNN architecture we used is shown in Figure 2, which is similar as the architecture first proposed by Alex[14]. The architecture takes images of the size 256*256*3 as input. Since we generate image translations and horizontal reflections before training, the size of the input images is 224*224*3. The architecture has four convolutional layers and the first and the second layers are followed by max-pooling layers and normalization layers. We can also see in Figure 2 that the number of filters of the four convolutional layers is 64,96,96 and 64 respectively, and the kernel size of the four convolutional layers is 7*7,5*5,3*3 and 3*3 respectively.

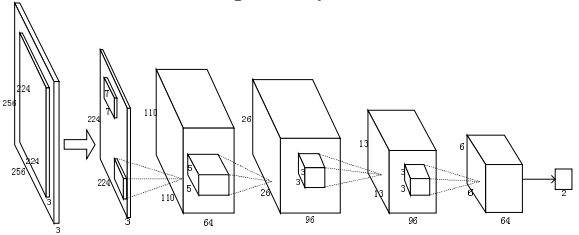

Figure 2. The traditional DCNN architecture we used for image aesthetic evaluation. The architecture has four convolutional layers and the first and the second layers are followed by max-pooling layers and normalization layers.

We can know from [14] that the architecture of DCNN can significantly affect the performance. Therefore it is necessary to evaluate the performance of architectures with different number of convolutional layers. To determine the architecture of the traditional DCNN, we conducted experiments on the candidate architectures shown in Table 1 on the category of Architecture in PhotoQualityDataset[6] and finally picked the one with the best performance. We define the DCNN architecture with 3 convolutional layers, 4 convolutional layers and 5 convolutional layers as Arch1, Arch2, and Arch3 respectively. Test error rates achieved by the three architectures are also given in Table 1. We can see from Table 1 that the performance of Arch2 with 4 convolutional layers is the best.

Table 1 Test error rates of DCNN architectures with 3,4,and 5 convolutional layers respectively on category of Architecture in PhotoQualityDataset

| Architectures | conv1 | pool1 | rnorm1 | conv2 | pool2 | rnorm2 | conv3 | pool3 | rnorm3 | conv4 | conv5 | fc2 | error rates |
|---|---|---|---|---|---|---|---|---|---|---|---|---|---|
| Arch1(3 layers) | √ | √ | √ | √ | √ | √ | √ | √ | √ | | | | 0.09916 |
| Arch2(4 layers) | √ | √ | √ | √ | √ | √ | √ | √ | √ | √ | | √ | **0.08571** |
| Arch3(5 layers) | √ | √ | √ | √ | √ | √ | √ | √ | √ | √ | √ | √ | 0.09832 |

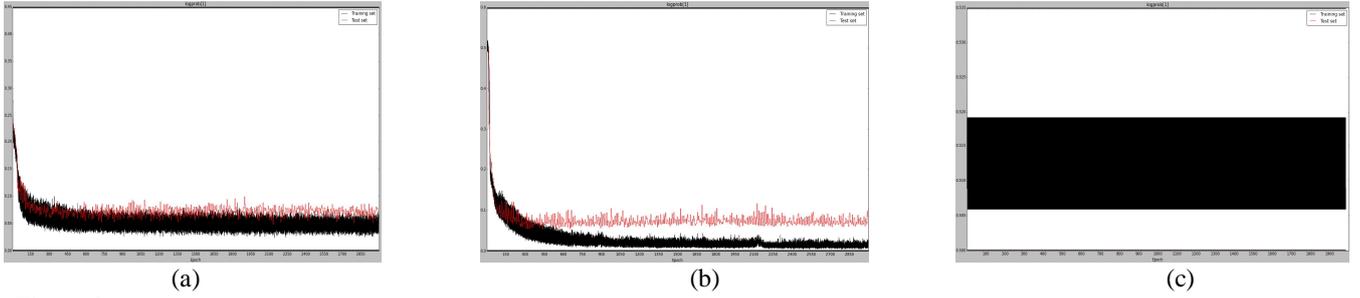

(a)            (b)            (c)

Figure 3 Classification error rates vary with the number of training iterations for traditional DCNN architectures of different number of layers, namely two, four and five. (a) two layers; (b) four layers; (c) five layers.

### 3.2. FRAMEWORK OF OUR PARALLEL DCNN

The traditional DCNN architecture can achieve a competitive performance in image aesthetic evaluation, however this architecture may meet the over-fitting and under-fitting problems when we conduct the experiments on the category of Human in PhotoQualityDataset [6]. Figure 3 shows the curve of the classification error rates varies with iteration numbers for architectures with different layer numbers, namely two, four and five. In the architecture with two layers, training error rates stop falling at 150 epochs and keep stable after 150 epochs. It is clear that this architecture meets under-fitting problem. In the architecture with five layers, the gradient has been diffused, the training error rates stop at first, and keep stable. It means that different architectures of CNN may meet different problem when training a specific dataset.

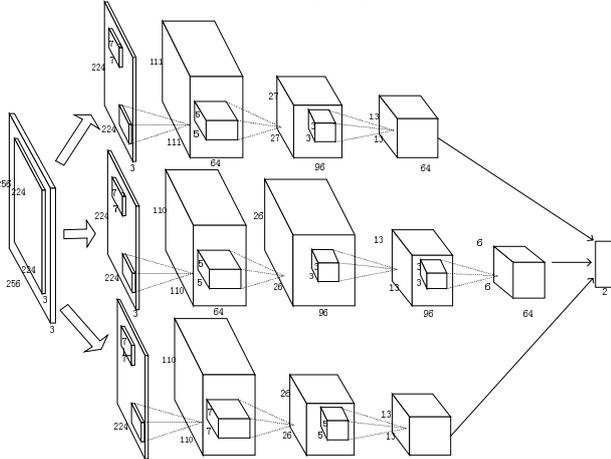

Figure 4. The 3-PDCNN architecture we proposed. It parallels three DCNNs.

In order to prevent under-fitting and over-fitting problems. We proposed the PDCNN architecture and the architecture of our PDCNN is given in Figure 3. To select a PDCNN architecture which performs the best in image aesthetic evaluation, we conducted experiments on the basis of Section 3.1. We conducted several experiments to select the best paralleled architecture. We defines the PDCNN which parallel $n$ DCNNs as $n$-PDCNN. For example, 2-PDCNN means a PDCNN architecture which parallel two DCNNs. There are various kinds of combinations of DCNN when selecting the best 2-PDCNN architecture, we fix the best architecture selected in Section 3.1 as one of the two DCNNs and parallel with other DCNN of 3 layers, 4 layers and 5 layers respectively. When selecting the best 3-PDCNN architecture, we fix the best 2-PDCNN architecture chosen before as two of the three DCNNs and parallel with other DCNN of 3 layers, 4 layers and 5 layers respectively. When selecting the best 4-PDCNN, we use the same method as selecting the best 3-PDCNN architecture. According to this principle, we conducted experiments on candidate architectures shown in Table 2, Table 3 and Table 4 on category of Architecture in PhotoQualityDataset and finally picked the one with the best performance respectively.

Table 2 Test error rates achieved by candidate 2-PDCNN architectures of category of architecture in PhotoQualityDataset

| Architecture | Paralleled DCNNs | Error rate |
|---|---|---|
| ParaArch1 | Arch2 | **0.082353** |
| | Arch1 | |
| ParaArch2 | Arch2 | 0.088235 |
| | Arch2 | |
| ParaArch3 | Arch2 | 0.107653 |
| | Arch3 | |

Table 3 Test error rates achieved by candidate 3-PDCNN architectures of category of architecture in PhotoQualityDataset

| Architecture | Paralleled DCNNs | Error rate |
|---|---|---|
| ParaArch4 | ParaArch1 | 0.081513 |
| | Arch1 | |
| ParaArch5 | ParaArch1 | **0.079832** |
| | Arch2 | |
| ParaArch6 | ParaArch1 | 0.089916 |
| | Arch3 | |

Table 4 Test error rates achieved by candidate 4-PDCNN architectures of category of architecture in PhotoQualityDataset

| Architecture | ParalleledDCNN | Error rate |
|---|---|---|
| ParaArch7 | ParaArch5 | 0.094118 |
| | Arch1 | |
| ParaArch8 | ParaArch5 | **0.083193** |

| | Arch2 | |
|---|---|---|
| ParaArch9 | ParaArch5 | 0.089916 |
| | Arch3 | |

We can see from Table 2 that the architecture ParaArch2 which paralleled a DCNN of four convolutional layers and a DCNN of three convolutional layers performs the best. It is mainly because that the PDCNN architecture with different architectures can learn more discriminative features. Results in Table 3 show that the PDCNN which paralleled two DCNNs of three convolutional layers and a DCNN of four convolutional layers performs the best, which is also better than architecture ParaArch2. It is noted that the kernel size of these two DCNNs of three convolutional layers is not the same, since we aim at learning more discriminative and different high-level features. From Table 4 we can see that the PDCNN paralleled with four DCNNs performs worse than that paralleled with two or three DCNNs, it is mainly because that it is too complicated to learn suitable image aesthetic features. Above all, we can conclude that the best number of paralleled DCNNs is three. In other words, the 3-PDCNN architecture with a DCNN of three layers and two DCNNs of four layers performs the best in image aesthetic evaluation.

## 3. EXPERIMENTS AND ANALYSIS

### 3.1. Datasets

In order to evaluate the performance of the proposed methods, experiments are conducted on two published image aesthetic evaluation datasets named PhotoQualityDataset [6] and CUHK dataset [4]

PhotoQualityDataset consists of images from a website which gathered images taken by both professional photographers and amateurs. According to the contents of images, all images are divided into 7 categories, named Animal, Architecture, Human, Landscape, Night, Plant and Static. All images are labeled by 10 independent reviewers into 3 classes: high quality, low quality and uncertain about quality. An image will be labeled as high quality or low quality if 8 out of 10 reviewers agree on the label given individually. The number of images of each category in PhotoQualityDataset is given in Table 5. It is apparently that PhotoQualityDataset is an unbalanced dataset, since the number of high quality images of each category is much smaller than that of low quality images The PhotoQualityDataset is too small to train CNN networks, therefore, it is necessary to produce more image data from the existing dataset. All images of each category in PhotoQualityDataset will be rotated to 90 degree, 180 degree and 270 degree respectively, and both the raw images and the rotated images are randomly divided into 4 data batches. Three batches are taken as the training set, and the remaining one is taken as the testing set.

CUHK dataset[4] consists of a diverse set of high and low quality images from an image contest website (www.DPChallenge.com). The obtained 60,000 images are rated by hundreds of users at DPChallenge.com. The top 10%, i.e. 6000 images in total, are rated as high quality images, and the bottom 10% are low quality images. We randomly choose 3000 high quality and 3000 low quality images as the training set, and take the remaining 3000 high quality and 3000 low quality images as the test set.

### 3.2. Training skills

Stochastic gradient descent is used to train our model with a mini-batch size of 32 images, a momentum of 0.9, and a weight decay of 0.0005. Moreover, in order to extend the datasets, we follow the idea in [14] and introduce a data augmentation strategy. The form of data augmentation consists of generating image translation and horizontal reflections, and then we randomly extract patches of the size of 224*224*3 from source images of the size of 256*256*3. By this way, our training dataset will be increased by a factor of 2048.

### 3.3. Experiments on PhotoQualityDataset

In order to compare the performance of the proposed PDCNN architecture with the traditional DCNN architecture without parallelization, we first conducted experiments on the traditional DCNN architecture, then we conducted experiments on the proposed PDCNN architecture. Recognition accuracies between our PDCNN and traditional DCNN are given in Table 6, and together with the recognition accuracies of six traditional feature extraction methods in[2, 3, 4, 6, 12,15].

Table 6 shows that the highest overall accuracy of seven categories is achieved by the proposed 3-PDCNN architecture, which is 1.5% higher than the best feature extraction method by Luo[6] and 0.7% higher than the proposed 2-PDCNN architecture. Compared with the DCNN method DCNN_Aesth_SP proposed by Zhe Dong [12] which combined spatial pyramid method with DCNN method, our 3-PDCNN architecture can also achieve 0.5% higher accuracy. We can see from Table 6 that the proposed 3-PDCNN performs the best in the categories of animal, architecture and landscape, with the performance of the other four categories closely to the best results. The proposed 2-PDCNN performs the best in the categories of human and night. The best result in the category of plant was achieved by Luo [6] since two kinds of specific features was designed for this category, named complexity combined feature and hue composition feature. The best result in the category of human was also achieved by Luo[6] since a face combined feature based on face detecting was used, which is able to achieve the accuracy of 95.21% simply by using face combined feature. In the category of night where images are always dark and the subject region is difficult to extract, the feature extraction methods in [2, 3, 4] based on subject region only achieved about 75% accuracy. Method in [6] achieved a much higher accuracy of 83.09% by designing the dark channel feature and scene composition feature which are able to represent the image aesthetics. Method in [15] achieves 87.42% accuracy because it uses semantic features to implicitly represent the image topic, which is helpful when the subject region extraction fails in the category of night. However, we can achieve a much higher accuracy than those traditional feature extraction methods of 87.96% simply by applying the traditional DCNN method to image aesthetic evaluation. Furthermore, the 2-PDCNN architecture can achieve 89.66% accuracy and 3-PDCNN architecture can achieve 88.39% accuracy, which is a great breakthrough in the category of night compared with traditional methods based on hand-crafted features. All the results indicate that the 3-PDCNN architecture can perfectly learn the image aesthetic features from the training images.

We can achieve much better performance than traditional methods simply by 3-PDCNN architecture or 2-PDCNN architecture. In order to get further better performance, we can choose the parallel architecture if we know the category of test images in advance. For example, we can choose the 3-PDCNN architecture when testing category of landscape and choose 2-PDCNN architecture when testing category of human. So we can get an overall accuracy of 92.64% when combining the performance of 3-PDCNN architecture and 2-PDCNN architecture, which is the best results ever. Therefore, we can efficiently achieve better performance in image aesthetic evaluation by paralleling DCNNs and features learnt by PDCNN architectures are more powerful than those hand-crafted features.

**3.4 Experiments on CUHK dataset**

The number of images in CUHK dataset is much larger than that in each category of the PhotoQualityDataset, however, our PDCNN architecture on the CUHK dataset is the same as that applied on PhotoQualityDataset. It is because that the proposed paralleled architecture has fused DCNN with different complexity and it is able to adapt to training dataset with different sizes well. Comparison of test error rates between the 2-PDCNN and 3-PDCNN method and three traditional feature extraction methods[4,3,16] are given in Table 7.

Experimental results in Table 7 show that the traditional DCNN method which simply apply DCNN on raw image data can perform better than all the feature extraction methods, even some of them are composed of several kinds of features, such as hand-crafted feature extraction method in[15]. Furthermore, the proposed 2-PDCNN architecture achieves lower test error rate than the traditional DCNN method, with a decrease by 1%. Compared with the proposed 2-PCDNN architecture, the proposed 3-PDCNN architecture performs even better, with a decrease by about 3%. Though the number of images in the CUHK dataset is much larger than each category in PhotoQualityDataset, we can achieve good performance with a PDCNN architecture which shares the same layer number, the same filter number and the same kernel size with the architecture used on the PhotoQualityDataset, which indicates that our PDCNN architecture is able to adapt to datasets in different sizes.

Table 5 Number of high quality and low quality photos of seven categories in PhotoQualityDataset

| Category | Animal | Architecture | Human | Landscape | Night | Plant | Static |
|---|---|---|---|---|---|---|---|
| Number of high quality photos | 947 | 595 | 678 | 820 | 352 | 594 | 531 |
| Number of low quality photos | 2224 | 1290 | 2536 | 1947 | 1352 | 1803 | 2004 |

Table 6 Comparison of average accuracies of each category in the PhotoQualityDataset using different methods

| Category | Animal | Plant | Static | Architecture | Landscape | Human | Night | Overall |
|---|---|---|---|---|---|---|---|---|
| Features proposed by Datta [2] | 0.7861 | 0.7638 | 0.7174 | 0.7386 | 0.7753 | 0.7694 | 0.6421 | 0.7495 |
| Features proposed by Ke[3] | 0.7751 | 0.8093 | 0.7829 | 0.8526 | 0.8170 | 0.7908 | 0.7321 | 0.7944 |
| Features proposed by Luo [4] | 0.8161 | 0.8238 | 0.8174 | 0.7386 | 0.7753 | 0.7794 | 0.6421 | 0.7792 |
| Hand-crafting feature proposed by Guo[15] | 0.8755 | 0.8936 | 0.9033 | 0.8509 | 0.8766 | 0.9219 | 0.8806 | 0.8884 |
| Semantic feature proposed by Guo[15] | 0.8623 | 0.8685 | 0.8964 | 0.8644 | 0.8416 | 0.9313 | 0.8742 | 0.8787 |
| Features proposed by Luo [6] | 0.8712 | **0.9147** | 0.8890 | 0.9004 | 0.9273 | **0.9631** | 0.8309 | 0.9044 |
| DCNN_Aesth_SP method by [12] | - | - | - | - | - | - | - | 0.9193 |
| Traditional DCNN architecture | 0.9182 | 0.8778 | 0.8766 | 0.8992 | 0.9244 | 0.9558 | 0.8796 | 0.9088 |
| The proposed 2-PDCNN architecture | 0.9307 | 0.8956 | 0.8936 | 0.9176 | 0.9366 | 0.9432 | **0.8966** | 0.9191 |
| **The proposed 3-PDCNN architecture** | **0.9323** | 0.8867 | 0.8804 | **0.9202** | **0.9500** | 0.9558 | 0.8839 | 0.9198 |
| **The proposed 2-PDCNN and 3-PDCNN combined** | 0.9323 | 0.8956 | 0.8936 | 0.9202 | 0.9500 | 0.9683 | 0.8966 | **0.9264** |

Table 7 Comparison of test error rates on the CUHK dataset using different methods

| Methods | Features proposed in [4] | Semantic feature proposed in [15] | FV-SIFT-SP proposed in [11] | Hand-crafting feature proposed in][15] | Traditional DCNN method | The proposed 2-PDCNN method | The proposed 3-PDCNN method |
|---|---|---|---|---|---|---|---|
| Test error rates | 0.2417 | 0.2247 | 0.1787 | 0.1717 | 0.1691 | 0.1592 | 0.1565 |

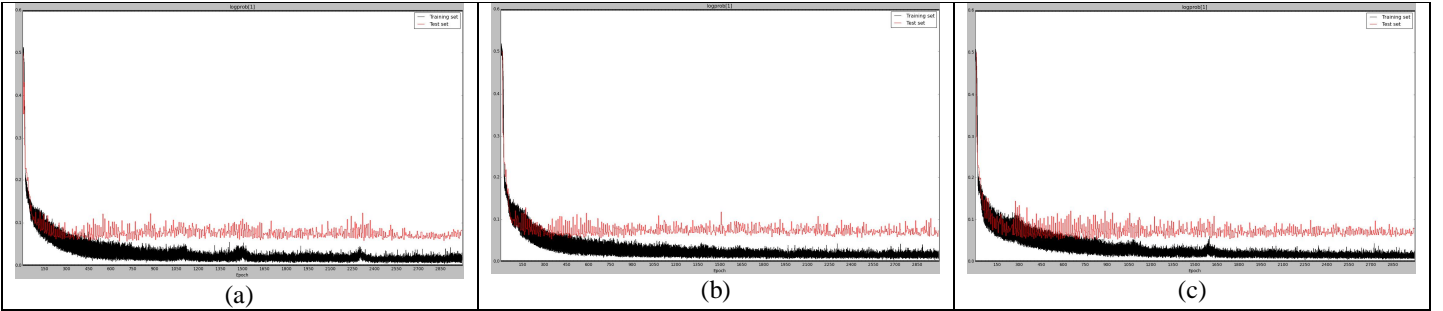

Figure 5 Classification error rates vary with the number of training iterations for paralleled architectures of different number of DCNNs, namely two, three and four. (a) two paralleled DCNNs; (b) three paralleled DCNNs; (c) four paralleled DCNNs.

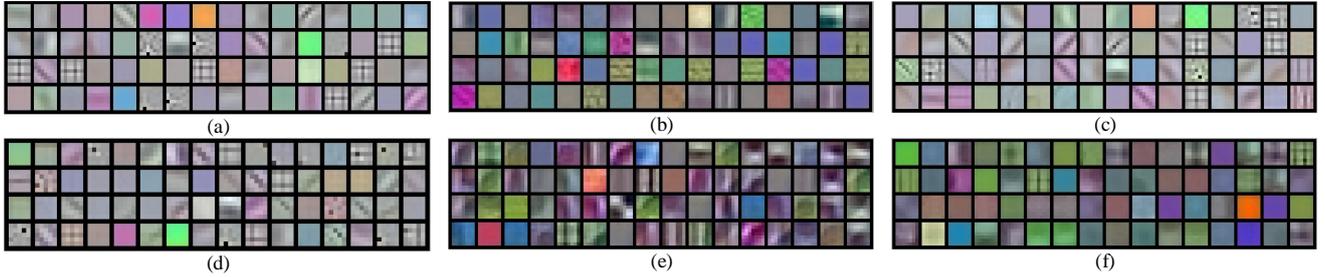

Figure 6 Filters of the first convolutional layers of Arch2, ParaArch1 and ParaArch5 respectively. (a) Filters of conv1 of architecture Arch2; (b) Filters of conv1a of architecture ParaArch1; (c) Filters of conv1b of architecture ParaArch1; (d) Filters of conv1a of architecture ParaArch5; (e) Filters of conv1b of architecture ParaArch5; (f) Filters of conv1c of architecture ParaArch5;

### 3.5 Analysis on under-fitting and over-fitting problem

In order to prove that our proposed PDCNN architectures are able to prevent under-fitting and over-fitting problems, we give the curve of classification error rates vary with the number of training iterations for PDCNN architectures of different number of DCNNs, namely two, three and four, as shown in Figure 4. We can see that no matter how many DCNNs are paralleled, the test error rates decline during a period time and then keep stable, which proves that it is hard to meet under-fitting and over-fitting problems.

In addition to the curve of classification error rates vary with the number of training iterations, Figure 6 shows the filters of the first convolutional layer of architecture Arch2 (shown as (a)), the two filters of the first convolutional layer of architecture ParaArch1 (shown as (b), (c)), the three filters of the first convolutional layer of architecture ParaArch5 (shown as (d), (e), (f)) respectively to better visualize the filters learnt. Generally speaking, good learnt filters are distributed regularly and are also obviously contrastive. It is apparent that filters learnt by architecture ParaArch1 are better than that learnt by architecture Arch2, and filters learnt by architecture ParaArch5 are better than that learnt by architecture ParaArch1. To better illustrate our view, we take the variance of the first convolutional layer as the metric of whether filters are good learnt. The higher variance, the better filters. Table 8 shows the variance of the first convolutional layer of architecture Arch2, ParaArch1, ParaArch5 respectively.

Table 8 Variances of the first convolutional layer of architecture Arch2, ParaArch1, ParaArch5 respectively

| Architecture | Arch1 | ParaArch1 | | ParaArch5 | | |
|---|---|---|---|---|---|---|
| | conv1 | conv1a | conv1b | conv1a | conv1b | conv1c |
| Variance | 0.007642 | 0.013350 | 0.006873 | 0.007687 | 0.020600 | 0.013348 |

We can see from Table 8 that the mean variance of architecture ParaArch5 is the highest, which indicates that the filters learnt by architecture ParaArch5 are the best. Though variance of conv1a of architecture ParaArch1 is a little bit lower than that of architecture Arch1, the variance of conv1b of architecture ParaArch1 is much higher. It indicates that architecture ParaArch1 can learn better filters of higher variance than architecture Arch1. So we can conclude that paralleled architectures are better in learning filters than that none-paralleled architectures and the PDCNN architecture with three DCNNs is the best in learning filters.

### 3.6 Computational efficiency

We conducted experiments on one Geforce Titan GPU. Comparison of the convergence time of the category of Architecture in PhotoQualityDataset spent by the best traditional DCNN architecture, 2-PDCNN architecture, 3-PDCNN architecture respectively is given in Table 9. We define $t$ as the mean training time that one batch spends in an iteration, $n$ as the number of training batches, $e$ as the convergence epochs and $T$ as the total convergence time. We can see from Table 9 that the 3-PDCNN architecture spends the least time of 17359 seconds (about 4.822 hours) to convergence. The longest time is 31959 seconds (about 8.875 hours), which is also very fast.

Table 9 Comparison of the convergence time of the category of Architecture in PhotoQualityDataset spent by the best traditional DCNN architecture, 2-PDCNN architecture, 3-PDCNN architecture respectively

| Architectures | $t$ /second | $n$ | $e$ | $(T = t * n * e)$ /second |
|---|---|---|---|---|
| DCNN | 8.32633 | 3 | 967 | 24155 |
| 2-PDCNN | 10.78233 | 3 | 988 | 31959 |
| 3-PDCNN | 6.26900 | 3 | 923 | 17359 |

## 4. CONCLUSION

In addition to applying DCNN to image aesthetic evaluation area to solve difficulties in designing and extracting better hand-crafted features of traditional methods, we also propose a PDCNN architecture to enhance the generality ability of DCNN in fitting datasets of different sizes. Experiments are conducted on two acknowledged datasets, named the PhotoQualityDataset and the CUHK dataset. Experimental results show that our PDCNN architecture is able to achieve better performance than the traditional DCNN architecture and performs much better than all the traditional feature extraction methods. Detailed analysis on filters learnt and the curve of classification error rates vary with the training iterations also shows that our PDCNN architecture can overcome under-fitting and over-fitting problems. In the future, we will explore other possible parallel algorithms.